\documentclass[mlabstract,final]{jmlr}

\setcitestyle{round}

\usepackage[utf8]{inputenc}
\usepackage[T1]{fontenc}
\usepackage{aas_macros}
\usepackage{wrapfig}
\usepackage{subcaption}
\usepackage{stmaryrd}
\usepackage{tikz}
\usepackage{standalone}
\usepackage{booktabs}
\usepackage[disable]{todonotes}   
\usetikzlibrary{positioning,arrows.meta,fit,backgrounds,calc,
                decorations.pathmorphing,decorations.shapes,shapes.geometric}

\definecolor{shared}{RGB}{33,102,172}
\definecolor{resid} {RGB}{27,150,118}
\definecolor{aux}   {RGB}{136,65,157}
\definecolor{frozen}{RGB}{90,90,100}
\definecolor{latent}{RGB}{214,158,30}
\definecolor{lossc} {RGB}{200,40,55}

\hypersetup{colorlinks=false}          
\usepackage[capitalize,noabbrev]{cleveref}

\usepackage[most]{tcolorbox}
\newtcolorbox{conclusionbox}{colback=blue!5,colframe=blue!75!black,
  coltitle=black,fonttitle=\bfseries,boxrule=1pt,arc=1mm,
  left=2mm,right=2mm,top=1mm,bottom=1mm}

\jmlrproceedings{}{NeurReps Workshop at NeurIPS, Extended Abstract Track}
\jmlrvolume{}
\jmlrpages{}
\firstpageno{1}
\jmlryear{2026}
\jmlrworkshop{Symmetry and Geometry in Neural Representations}

\title[Inducing Fourier Geometry]{Inducing Fourier Geometry: A Smoothness
Prior Imposes Approximate Translation Equivariance in Multimodal Stellar
Embeddings}

\author{
  \Name{Ilay Kamai} 
  \addr Technion - Israel Institute of Technology
  \AND
  \Name{Alex M. Bronstein}
  \addr ISTA - Institute of Science and Technology Austria, Technion - Israel Institute of Technology
  \AND
  \Name{Hagai B. Perets} 
  \addr Technion - Israel Institute of Technology 
  }
\begin{document}
\maketitle
\begin{abstract}
Learned representations often contain geometric structure induced by symmetries
in the underlying data. We ask whether a known continuous parameter space can
instead be used to \emph{induce} such structure when the corresponding symmetry
is only approximate. We introduce a graph-Laplacian smoothness prior that
encourages representations to vary smoothly over a physical parameter space
without specifying a representation geometry; in the continuum limit it selects
Laplacian eigenfunctions and therefore predicts the Fourier-like geometry
associated with translation symmetry. On multimodal stellar representations
spanning six astronomical surveys, the prior induces Toeplitz Gram structure and
sinusoidal modes along stellar parameters, including in modalities whose native
representations lack this geometry, while modality-specific residuals preserve
structure outside the imposed coordinates. Of the exact-symmetry predictions,
stationarity in a recoverable warped coordinate survives; diagnostics assuming
exact quantization or asymptotic rates do not. The prior also improves
downstream linear-probe performance and label efficiency, suggesting a general
strategy: where a meaningful continuous parameter space exists, its local
geometry can serve as a data-informed prior.
\end{abstract}
\section{Introduction}
\label{sec:intro}

Recent work on representation geometry has shown that translation-symmetric
structure in data statistics leaves a characteristic fingerprint in learned
embeddings: when pairwise co-occurrence statistics are approximately invariant
under shifts of an underlying coordinate, the representation Gram matrix is
approximately Toeplitz, its eigenvectors are Fourier modes, and one-dimensional
manifolds embed as Lissajous-like curves in successive principal components
\citep{Karkada2026_symmetry}. There the geometry \emph{emerges}, discovered by
generic objectives on data whose statistics happen to carry the symmetry. We ask
the converse: can it be deliberately \emph{induced} where the symmetry is only
approximate, and is it useful to do so?

Scientific data provides a natural testbed. Stellar properties vary continuously
with physical coordinates such as effective temperature ($T_{\mathrm{eff}}$) and
surface gravity ($\log g$), making translation along them physically meaningful,
yet stellar populations are not exactly stationary and different surveys carry
distinct noise, resolution, and selection effects. We study the shared
representation of a multimodal stellar embedding model in which six surveys are
aligned into a common latent space, augmenting the alignment objective with a
graph-Laplacian smoothness penalty in physical-label space. The penalty pulls
nearby stars in $(T_{\mathrm{eff}},\log g)$ together but specifies no Fourier
geometry; in the continuum limit it becomes a Dirichlet energy whose
fixed-variance minimizers are Laplacian eigenfunctions. Under approximate
stationarity, the Fourier geometry of \citet{Karkada2026_symmetry} is therefore
a \emph{prediction} of the prior rather than an empirical surprise.

The prior is not new in isolation; what differs is where we point it. A
graph-Laplacian Dirichlet penalty is manifold regularization
\citep{Belkin2006_manifoldreg}, with the density-weighted Laplacian of diffusion
maps as its continuum limit \citep{Coifman2006_diffusionmaps}, and the formalism
has been used for data fusion and alignment
\citep{Lafon_datafusion,Ham2005_manifoldalignment,Wang_alignemnt}. Here the
manifold is supplied externally, as a trusted physical coordinate space, and the
regularized object is a shared representation in a deep network. That only part
of a multimodal representation admits a common basis is long established
\citep{Tucker1958AnIM,Flury1984_commonPrinciples,Eynard2015_jointdiag,Klami2013_factorized,Tsai2018_mfm}
and parallels our shared/residual split; recovering a withheld modality can also
be seen as graph-regularized matrix completion
\citep{Kalofolias2014_mcgraphs,Boyarski2022_sgmc}.

Together our results turn Fourier-like representation geometry from a diagnostic
of symmetric data into a design target: a smoothness prior defined in
physical-label space installs Toeplitz Gram structure and sinusoidal modes in all
six surveys' shared representations, including three whose raw features lack
them, and improves linear probing on every label we test.
\section{Setting: data, model, and the smoothness prior}
\label{sec:setting}

\textbf{Data.} We use six large-scale stellar surveys spanning spectroscopy
(LAMOST, APOGEE, GALAH), spectrophotometry (\emph{Gaia} XP), and photometric
time series (\emph{Kepler}, TESS), together with \emph{Gaia} astrometry,
cross-matched through the \emph{Gaia} DR3 source identifier. We probe geometry
along continuous stellar parameters, primarily $T_{\mathrm{eff}}$, $\log g$,
and [Fe/H]; survey statistics and preprocessing are in
\Cref{sec:preprocessing}.

\textbf{Model.} Each modality is mapped by a pretrained, \emph{frozen} encoder
to features $z_i$. A Perceiver-IO-style alignment module maps these into a
$d=64$ \emph{shared} stream, trained for cross-survey agreement, and a $d=192$
per-survey \emph{residual} stream that can retain modality-specific structure.
The shared stream is trained with VICReg \citep{Bardes2021_vicreg}; astrometry
enters as a conditioning vector.

\textbf{The smoothness prior.} We add a density-normalised graph-Laplacian
penalty in physical-label space. Let $\bar{s}_a$ be the mean shared token of
star $a$ and $c_a=(T_{\mathrm{eff}},\log g)_a$ its normalized label
coordinates. With a fixed-bandwidth Gaussian kernel
$w_{ab}=\exp(-\lVert c_a-c_b\rVert^2/2\varepsilon^2)$ and degrees
$q_a=\sum_b w_{ab}$,
\begin{equation}
\mathcal{L}_{\mathrm{smooth}}
= \frac{1}{d_s}\,
\frac{\sum_{(a,b)\in E}\tilde{w}_{ab}\,\lVert \bar{s}_a-\bar{s}_b\rVert^2}
     {\sum_{(a,b)\in E}\tilde{w}_{ab}},
\qquad
\tilde{w}_{ab}=\frac{w_{ab}}{q_a\,q_b},
\end{equation}
where $E$ retains edges with $\lVert c_a-c_b\rVert<3\varepsilon$. The
$1/(q_aq_b)$ normalisation is the $\alpha=1$ member of the Coifman--Lafon
family \citep{Coifman2006_diffusionmaps}, which makes the continuum limit the
\emph{unweighted} Dirichlet energy $\int\lVert\nabla_c s\rVert^2\,dc$ rather
than a density-weighted one and ensures the penalty is not dominated by the most densely populated
regions of the HR diagram (\Cref{fig:alpha0}); the VICReg variance term, evaluated under a
matching density-free measure, fixes its scale (\Cref{app:fourier}).
Minimising then selects the low-lying eigenfunctions of the ordinary Laplacian,
which for an approximately one-dimensional coordinate are sinusoids of
increasing node count.
The penalty encodes no Fourier structure explicitly: the
geometry of \citet{Karkada2026_symmetry} is a predicted minimiser of the loss
rather than a feature we impose by hand.

\textbf{Measurements and scope.} Geometry is measured from bin-mean embeddings
along individual coordinates (or coordinate pairs), on signature-consistent
main-sequence samples to avoid survey-availability effects; we quantify Toeplitz
structure by the fraction of off-diagonal Gram variance explained
($R^2_{\mathrm{Toep}}$), count sign changes in leading eigenmodes, and fit
kernels to Gram-diagonal decay. Two scope conditions apply: stellar parameters
enter parts of the training pipeline, so we claim induced consistency, not
emergence from unlabeled data; and the physical symmetry is only approximate, so
we test explicitly which exact-symmetry predictions survive. Throughout, ``with
$\mathcal{L}_{\mathrm{smooth}}$'' denotes the full density-normalised prior---the
$\alpha{=}1$ kernel together with the $1/q$-weighted variance and covariance
statistics; the ablation removes both.
\section{Results}
\label{sec:measurements}

\begin{figure}[!ht]
\centering
\begin{minipage}{0.55\linewidth}
\includegraphics[width=\linewidth]{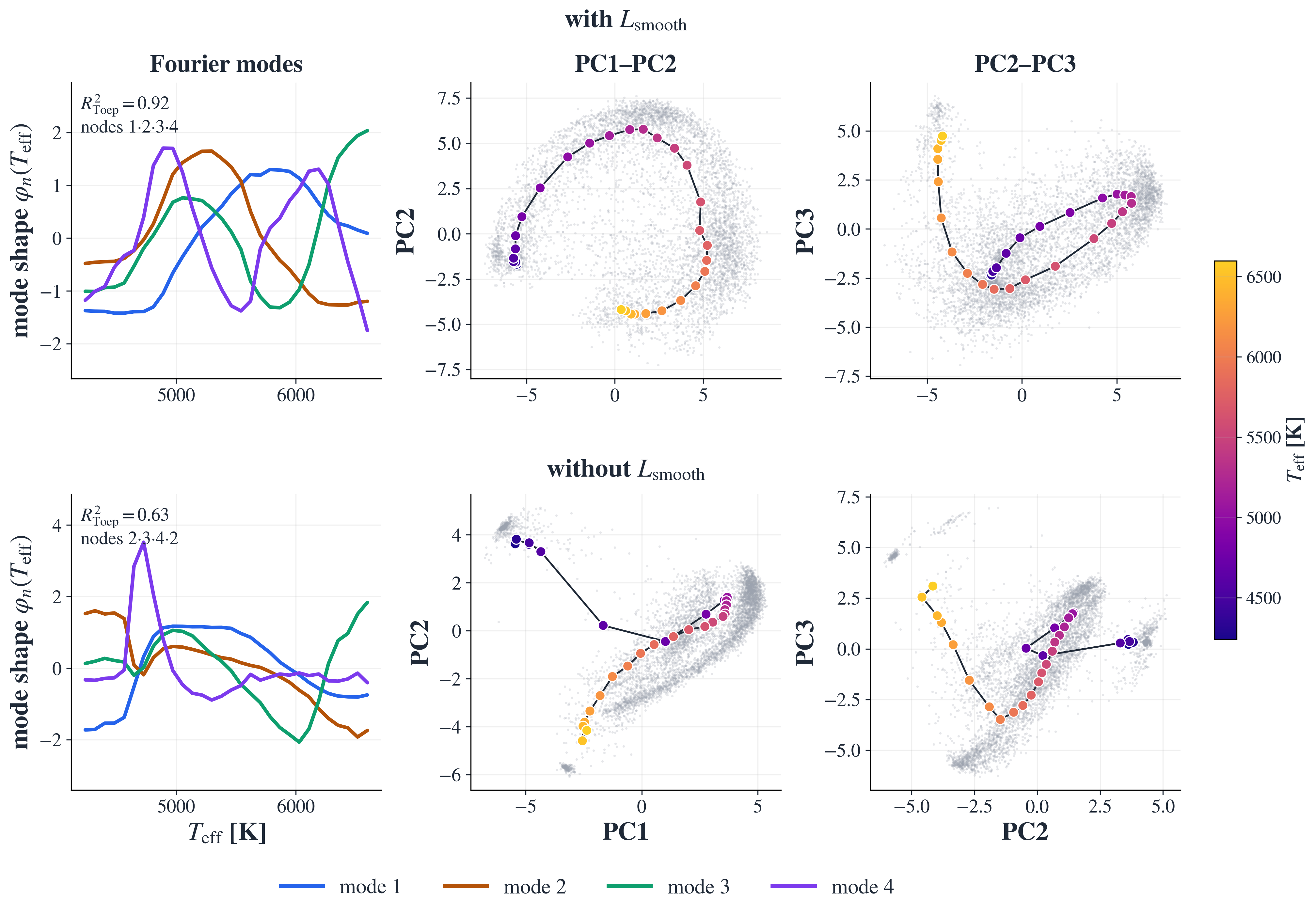}
\end{minipage}
\begin{minipage}{0.44\linewidth}
\caption{Translation symmetry in the shared latent space. \textbf{Upper
row:} our full model, with $\mathcal{L}_{\mathrm{smooth}}$. Fourier
modes in $T_{\mathrm{eff}}$ with frequency matching eigenmode degree,
and principal-component manifolds matching the prediction of
\citet{Karkada2026_symmetry}. \textbf{Lower row:} the same model
without $\mathcal{L}_{\mathrm{smooth}}$, showing neither.}
\label{fig:symmetry}
\label{fig:symmetry}
\end{minipage}
\end{figure}
 
\begin{figure}[!ht]
  \centering  \includegraphics[width=\linewidth]{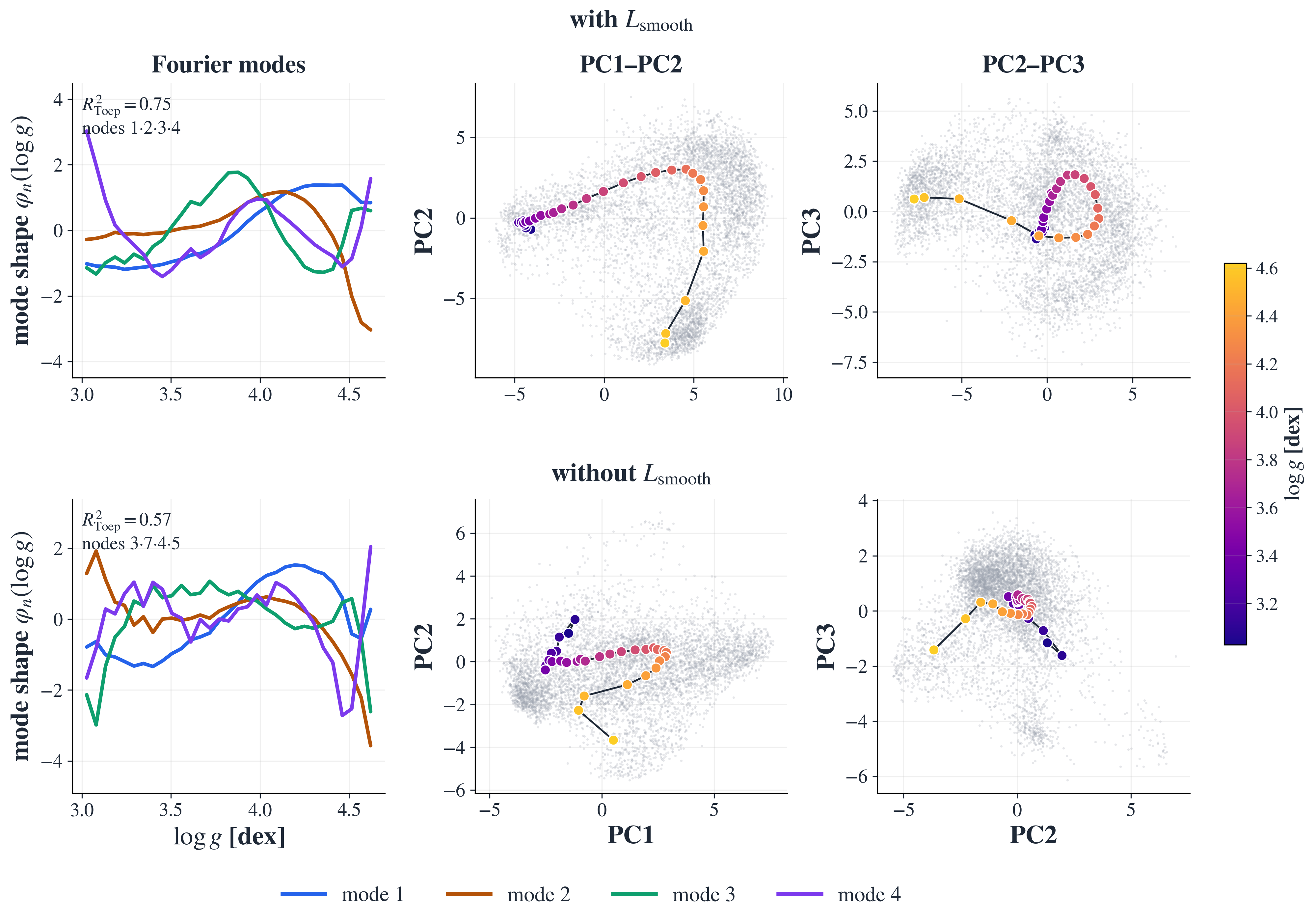}
  \caption{Same as~\Cref{fig:symmetry}, but for $\log g$.}
  \label{fig:symmetry_logg}
\end{figure}

\textbf{The prior installs the geometry.} The direct test compares the shared
stream with and without $\mathcal{L}_{\mathrm{smooth}}$, holding data,
architecture, and seed fixed. With the term, the modality-averaged shared
token's Gram matrix along $T_{\mathrm{eff}}$ is strongly Toeplitz
($R^2_{\mathrm{Toep}}=0.92$) with leading eigenmodes of node count
$1\cdot2\cdot3\cdot4$, tracing closed Lissajous curves in successive
principal-component pairs (\Cref{fig:symmetry}). Without it the fit drops to
$0.63$ and the node ladder disorders; $\log g$, the second graph coordinate,
behaves the same way (\Cref{fig:symmetry_logg}).

\begin{figure}[!ht]
  \centering  \includegraphics[width=\linewidth]{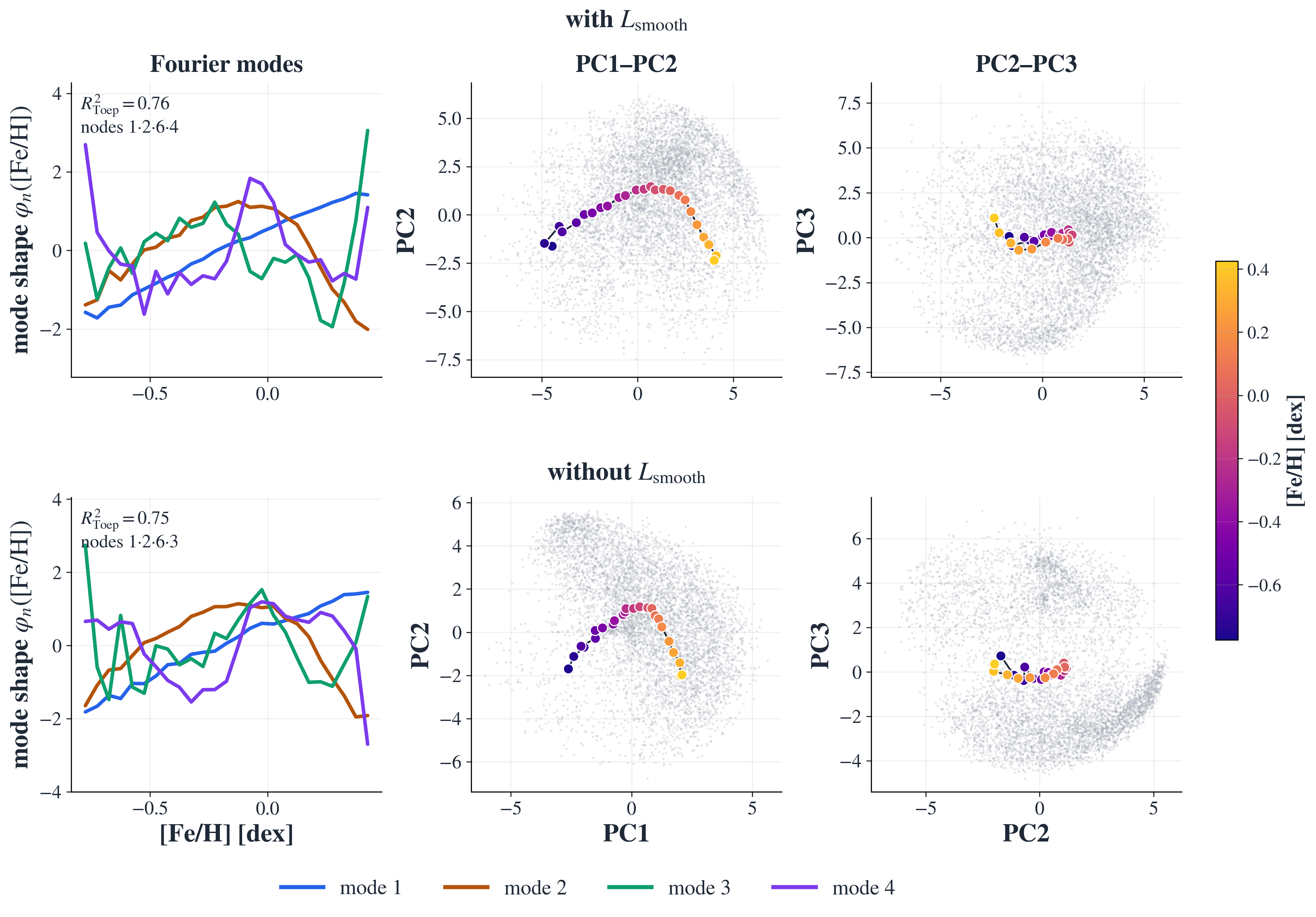}
  \caption{Same as~\Cref{fig:symmetry}, but for [Fe/H].}
  \label{fig:symmetry_feh}
\end{figure}

\begin{figure}[!ht]
    \centering
    \includegraphics[width=\linewidth]{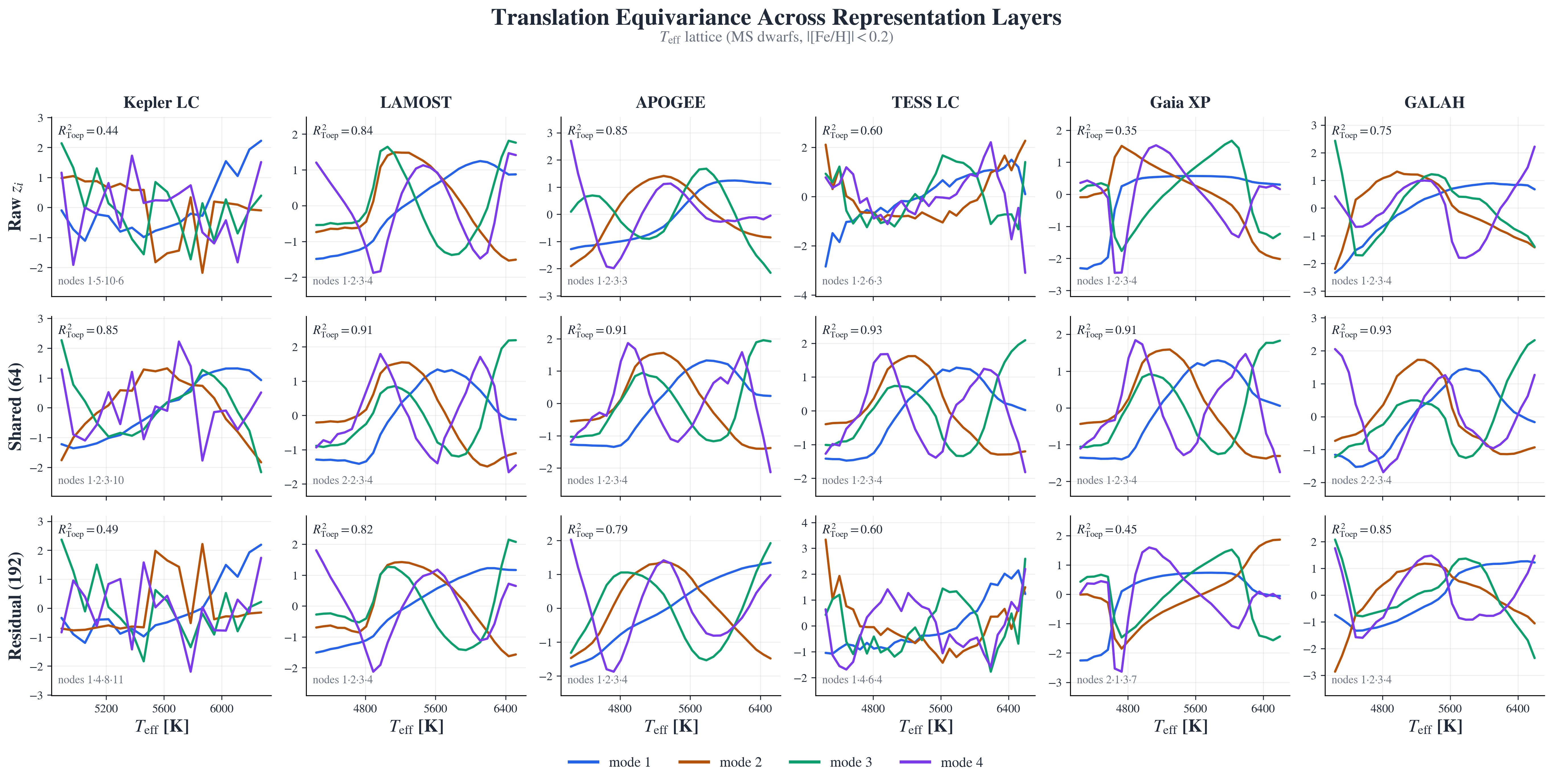
    }
    \caption{Fourier geometry per modality and layers in the model for $T_{\rm{eff}}$. Each column represents a specific modality embedding. Each row represents different parts in the model - raw pretrained features, $z_i$, (upper), shared space (middle), and residual space (lower). We see that the shared space improves Fourier geometry compared to $z_i$ across all modalities. The residual stream preserves the geometry of $z_i$.}
    \label{fig:layers}
\end{figure}

\begin{figure}[!ht]
    \centering
    \includegraphics[width=\linewidth]{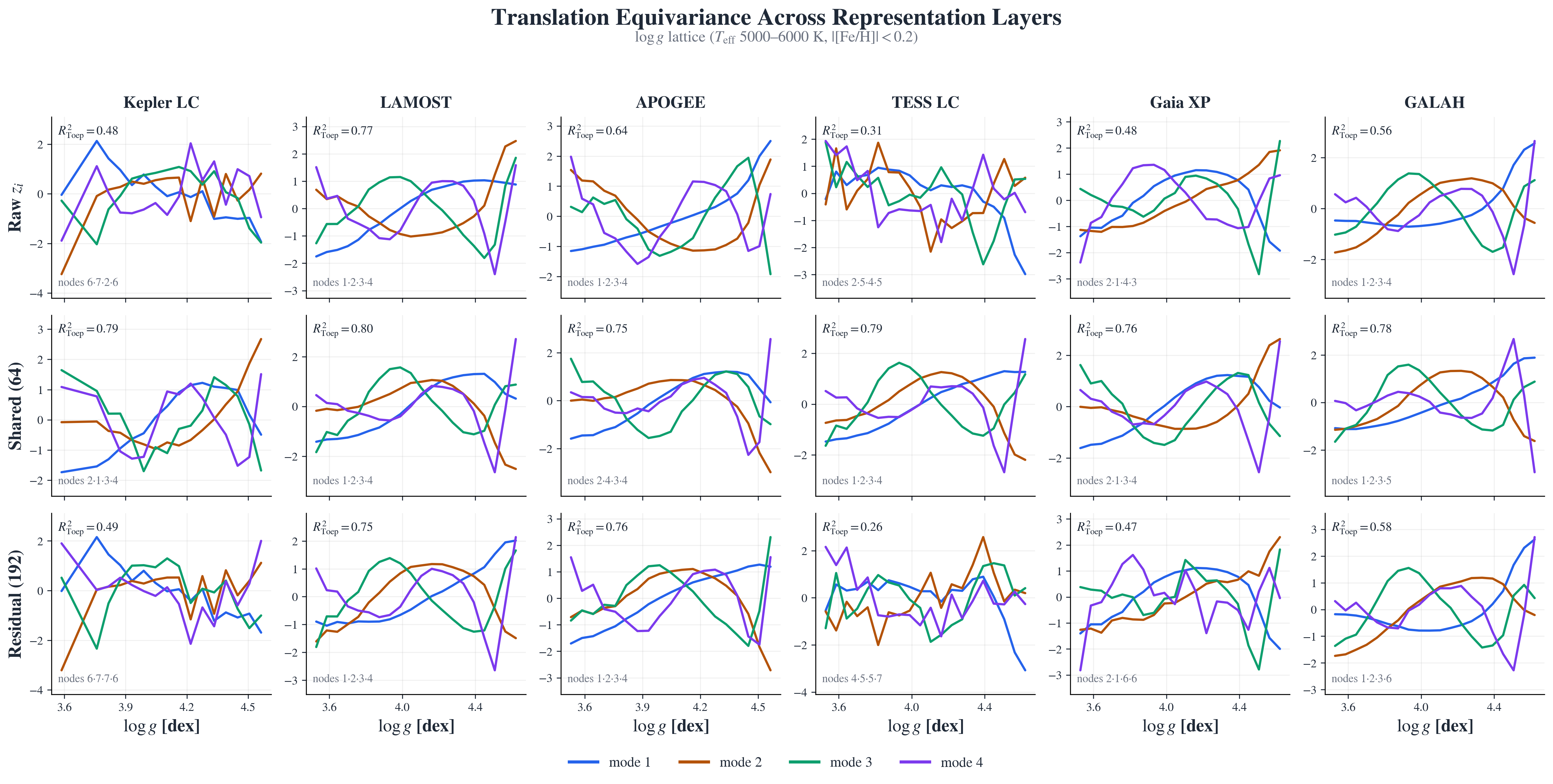
    }
    \caption{same as~\Cref{fig:layers}, for $\log g$}
    \label{fig:layers_logg}
\end{figure}

\begin{figure}[!ht]
    \centering
    \includegraphics[width=\linewidth]{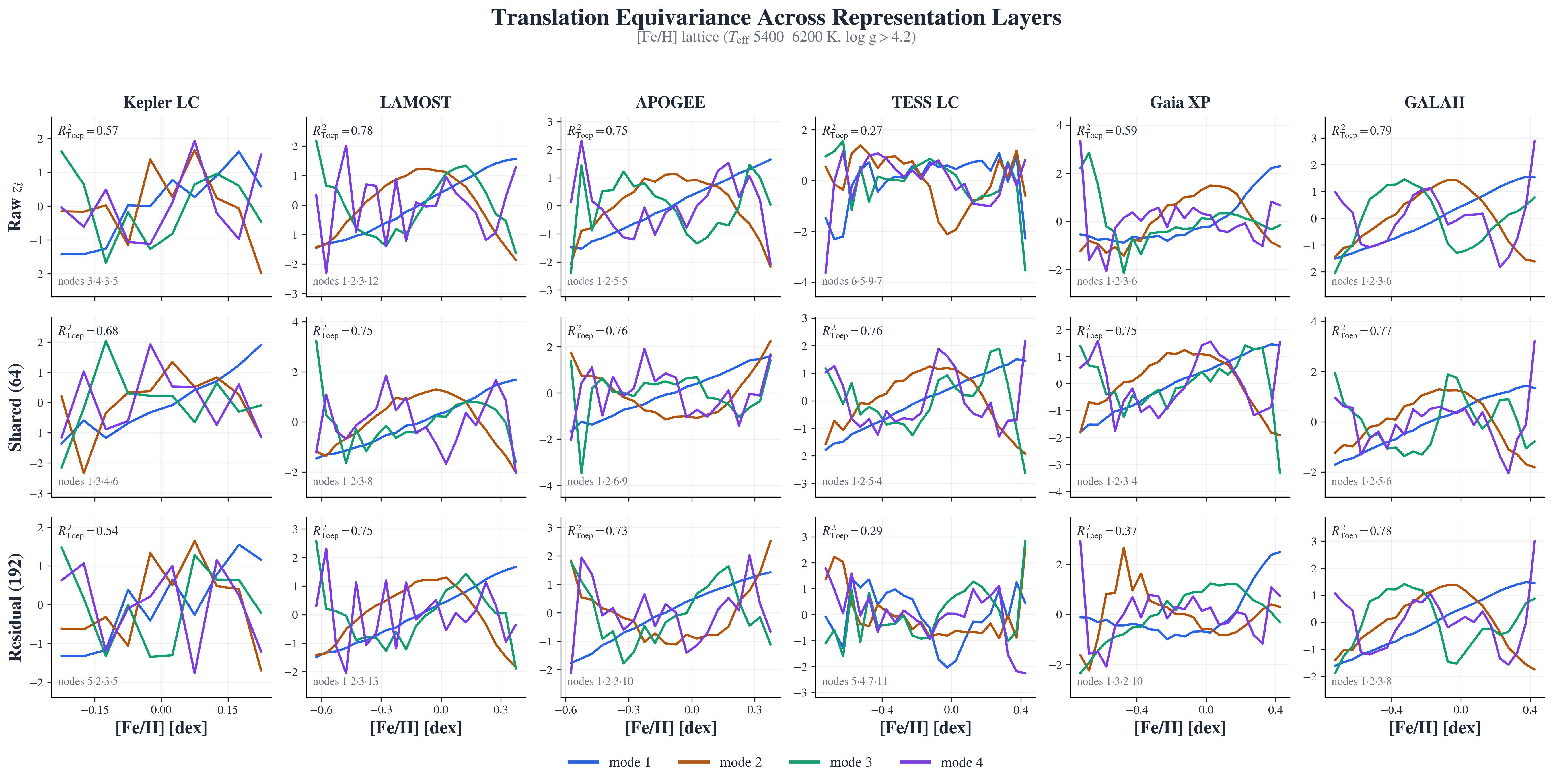
    }
    \caption{same as~\Cref{fig:layers}, for [Fe/H]. Note that this is not a graph coordinate.}
    \label{fig:layers_feh}
\end{figure}

\textbf{Where the geometry lives.} \Cref{fig:layers} probes three layers of the
with-prior model. In the raw features $z_i$ the symmetry is strong in the
spectroscopic surveys ($0.75$--$0.85$) and weak in the photometric and
spectrophotometric ones ($0.35$--$0.60$); the shared stream makes it stronger
and uniform (all six in $0.85$--$0.93$) while the residual stays uneven
($0.45$--$0.85$). The battery along $\log g$ reproduces the pattern
(\Cref{fig:layers_logg}). Along {[Fe/H]}, a coordinate the prior does not name,
the shared stream transfers the symmetry rather than taxing it: native
spectroscopic structure is approximately preserved while the surveys lacking it
gain it outright (TESS $0.27\to0.76$; \Cref{fig:layers_feh}). The transfer is
carried by alignment, not the prior: the pooled {[Fe/H]} Gram is
indistinguishable with and without $\mathcal{L}_{\mathrm{smooth}}$
($R^2_{\mathrm{Toep}}=0.76$ vs.\ $0.75$; \Cref{fig:symmetry_feh}). The induced
geometry is targeted: installed along the named coordinates, inert off-graph.

\textbf{Which symmetry diagnostics survive approximation.}
\citet{Karkada2026_symmetry} predict three signatures of a distance-only latent
coordinate: a Toeplitz Gram matrix, quantized eigenmode wavenumbers, and
decoding error $\sim r^{-1/D}$. Only the first survives, and Toeplitz structure
alone does not exclude a matched non-stationary kernel; the evidence is the
\emph{coordinate}: reparameterising by embedded arclength raises it to $0.97$,
and an independently fitted monotone warping agrees ($r=0.92$). Mode counting is
undecidable, its quantization ladder assuming an exponential kernel where our
spectrum is Gaussian ($\ell\approx440$~K), and decoding-error scaling deviates
($r^{-0.5}$ vs.\ $r^{-1}$ along $T_{\mathrm{eff}}$; $r^{-0.8}$ vs.\ $r^{-1/2}$
on the plane), consistent with a finite lattice below the asymptotic regime.

\textbf{Does the induced geometry pay?} Linear probes on the frozen shared
latent improve with the prior on every label at every training-set size
(\Cref{fig:probe_shared}), most where labels are scarce: age reaches
$R^2\approx0.38\pm0.09$ from $50$ labelled stars, comparable within error to
the without-prior probe at $20{,}000$. We attribute these gains to the prior as
a whole; whether Fourier structure specifically, rather than smoothness
generally, buys the improvement remains open.

\begin{figure}[!ht]
  \centering  \includegraphics[width=\linewidth]{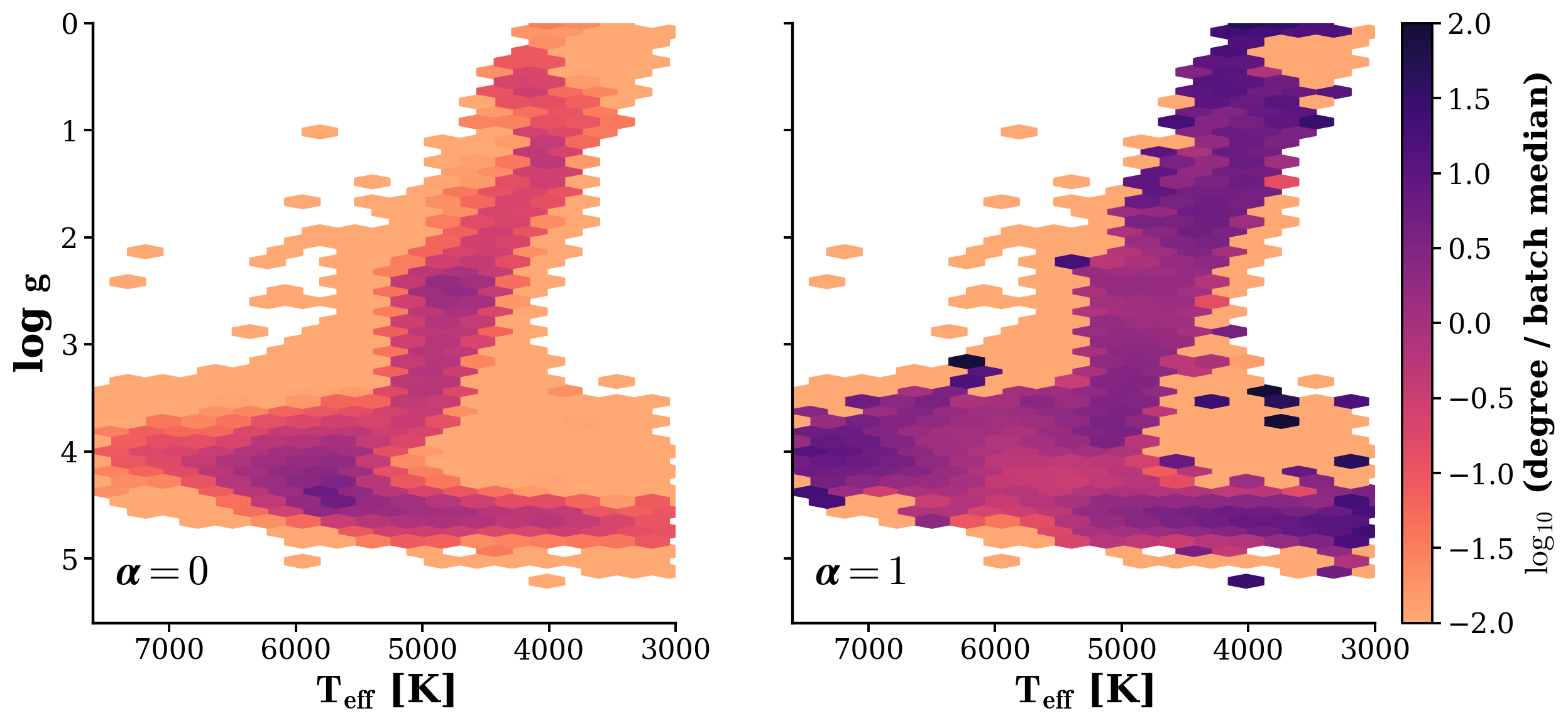}
  \caption{Where the smoothness penalty acts. Each star's graph degree, relative to the batch median ($\log_{10}$), binned in
$(T_{\mathrm{eff}}, \log g)$. \textbf{Left:} without the density
correction ($\alpha{=}0$), the weight concentrates on the densest
regions, the main sequence and red clump, and giants, hot stars and subgiants are nearly unweighted. \textbf{Right:} with it
($\alpha{=}1$, our model), the weight spreads across a wider  part of the HR diagram;
the few very high-degree bins at the sparse edges are stars the
correction up-weights where the density estimate is thin.}
  \label{fig:alpha0}
\end{figure}

\textbf{Physical coordinates, not generic smoothness.} We retrained the
model changing only the graph the Laplacian is built on, with data,
seed and recipe fixed (\Cref{tab:smoothctl}). A permutation null
(identical kernel, bandwidth and neighbour count, with
$(T_{\mathrm{eff}},\log g)$ shuffled across stars) falls below a
no-Laplacian baseline on every label and destroys the geometry
($R^2_{\mathrm{Toep}}=0.39$). A label-free regulariser built on the
model's own embedding also falls below it. Only the physical graph
improves on no-Laplacian across labels, by $+0.04$ to $+0.11$ on five
of six; a graph on the input embeddings, whose neighbours are only
partly physical, helps only on $T_{\mathrm{eff}}$ and $\log g$.
Without the density correction ($\alpha{=}0$), the penalty's weight
collapses onto the main sequence and red clump (\Cref{fig:alpha0}):
the geometry on our main-sequence lattice is the cleanest of any arm
($R^2_{\mathrm{Toep}}=0.91$), but probes over the full population are
weaker. A lattice score measured in one region therefore cannot rank
models that weight regions differently. What the controls do show is
that with the wrong neighbours, the geometry and the gain fail
together.

\textbf{Implied distribution.} A Fourier geometry further implies that leading
directions are arcsine-valued rather than Gaussian, the opposite of what
distributional priors such as SIGReg \citep{Balestriero2025_lejepa} target. This
holds in all six surveys with the prior, in four without it, and in none of a
vanilla VICReg latent (\Cref{app:arcsine}).

\begin{figure}[!ht]
    \centering
    \begin{minipage}{0.50\linewidth}
        \includegraphics[width=\linewidth]{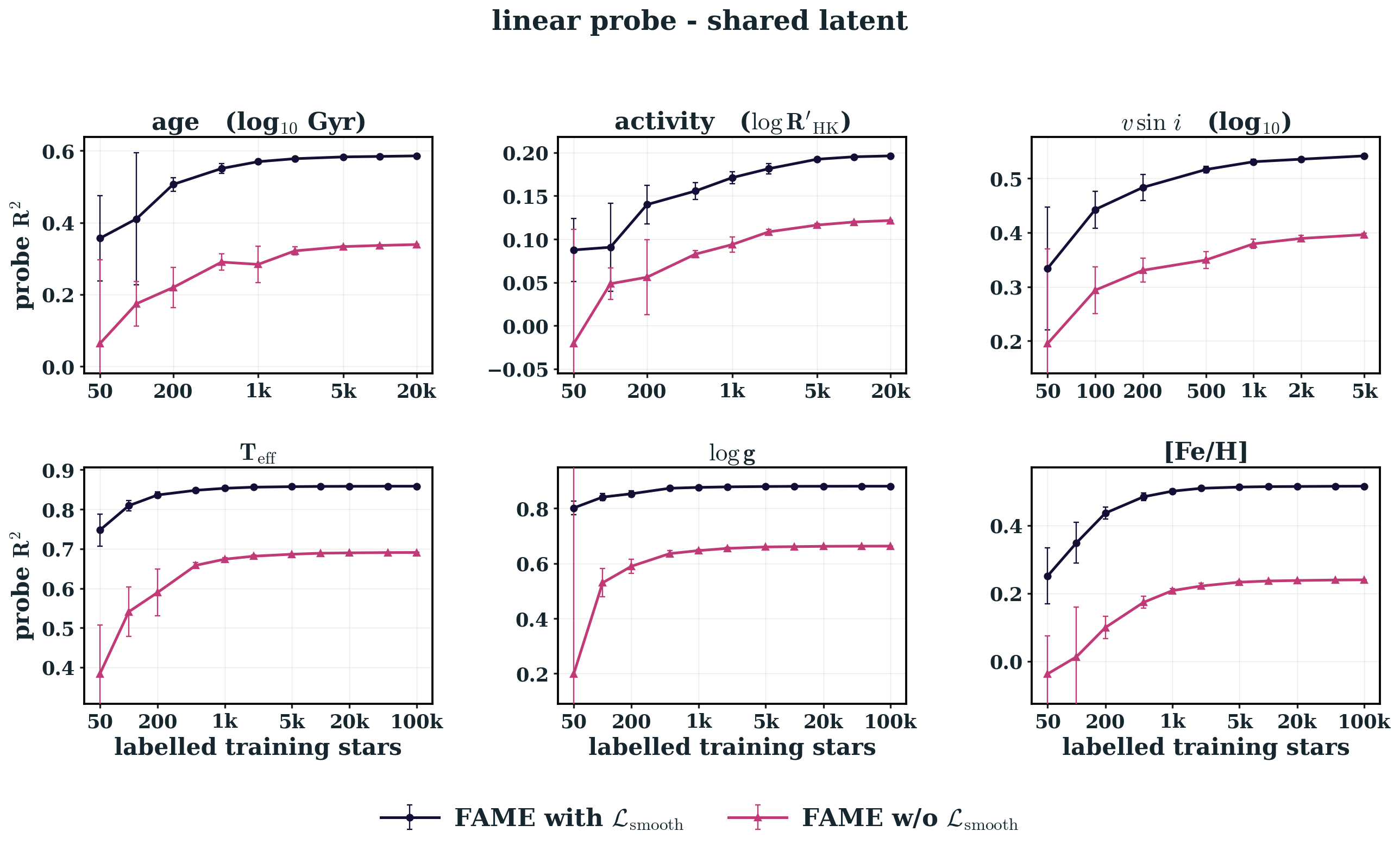}
    \end{minipage}
    \begin{minipage}{0.49\linewidth}
        \caption{Effect of $\mathcal{L}_{\mathrm{smooth}}$ on downstream
        tasks. Linear probes on the frozen shared latent, with
        $\mathcal{L}_{\mathrm{smooth}}$ (black) and without (magenta), against
        the number of labelled training stars.}
        \label{fig:probe_shared}
        \label{fig:probe_shared}
    \end{minipage}
\end{figure}
\section{Discussion and conclusions}
Emergence and induction are complementary: emergence reveals what data
statistics already contain, induction supplies what they do not. Here the
induction succeeds without any underlying physical invariance. What stellar
physics guarantees is that observables vary smoothly with $T_{\mathrm{eff}}$
and $\log g$, the same assumption behind spectral-grid interpolation, and
approximate stationarity follows where information content varies slowly along
the coordinate. The failures obey the same logic: age, whose governing physics
changes regime across the Kraft break, shows no symmetry at any layer, and
rotation only a partial one. The prior installs geometry where it is pointed
and nowhere else. Importantly, the prior consistently improves performance over all labels, not only on graph coordinates. This highlights the importance of such priors in scientific domains.

Two questions follow. Our controls rule out generic smoothness, but not
smoothness in physical coordinates: whether the Fourier structure itself
drives the downstream gains, or comes along with them, remains open, since
the two fail together when the graph's neighbours are wrong. And what
replaces exact quantization and decoding-rate predictions under approximate
symmetry is, to our knowledge, open. The physical sciences may be the
natural place to settle both, since there the coordinates are real and the
geometry can be checked against nature.

\bibliography{main}
\section{appendix}\label{sec:appendix}

\subsection{data and pre processing}\label{sec:preprocessing}
\paragraph{\textit{Kepler}.} The \textit{Kepler} mission
\citep{kepler:mission} monitored a single $\sim$115\,deg$^2$ field in Cygnus--Lyra
for four years at $30$-minute (long) cadence, delivering the highest-precision,
longest-baseline optical photometry available for main-sequence stars. We use
the long-cadence PDCSAP light curves (\cite{kepler:dr}) on their native,
uniform $48$\,samples\,day$^{-1}$ grid, with gaps zero-filled and no rebinning,
cut into $450$-day windows ($21{,}600$ points). Each window is lightly smoothed
(Savitzky--Golay, $\sim$0.5\,d) and expanded into six channels: standardized
flux, median-normalized flux, two activity-amplitude envelopes (a slow envelope
on a $6\,P_{\mathrm{rot}}$ window and a high-passed one on a
$1\,P_{\mathrm{rot}}$ window), the autocorrelation function (ACF, \cite{McQuillan2013_acf}), and the
Lomb-Scargle periodogram (LS, \cite{Lomb1976}, \cite{Scargle1982}). The resulting sample is $193{,}951$ stars.

\paragraph{TESS.} TESS \citep{tess:mission} trades \textit{Kepler}'s depth for
sky coverage, observing nearly the full sky in $\sim$27-day sectors at
multiple cadences. We use light curves from the
Quick-Look Pipeline \citep{tess:qlp}, restricted to $T_{\mathrm{mag}} < 16$ and
to targets observed in more than one sector. Each light curve is resampled to a
$30$-minute grid and truncated to $1000$ points in four channels: standardized
flux, median-filtered flux, ACF, and LS. These are the same channels as in \emph{Kepler}, but without the activity envelopes, as the sectors are too short for it. Compared to \emph{Kepler}, TESS
has a wider coverage
($1{,}503{,}260$ encoded stars) but has a shorter per-sector baseline, bigger pixel size ($21''$ vs. $4''$), and is generally noisier (\cite{VanCleve2016_tess_noise}.

\paragraph{LAMOST.} LAMOST \citep{lamost:survey} is a $4$-m telescope with $4000$ fibers, conducting the largest optical
spectroscopic survey of the northern sky. We use low-resolution
($R\!\sim\!1800$) spectra from \cite{lamost:dr}, continuum-normalized and
resampled onto a common $4096$-point wavelength grid. Its combination of scale
($5.1$M spectra) and northern footprint makes LAMOST the dominant spectroscopic
partner for both \textit{Kepler} and TESS.

\paragraph{APOGEE.} APOGEE \citep{apogee:mission} is a high-resolution
($R\!\sim\!22{,}500$) $H$-band spectrograph. Working in the near-infrared lets
it observe through dust into the Galactic plane and bulge, and it targets
predominantly giants---a population complementary to the dwarf-dominated
\textit{Kepler} and GALAH samples. We use the continuum-normalized combined
spectra from \cite{apogee:dr}, resampled to $8576$ points.

\paragraph{GALAH.} GALAH \citep{galah:mission} is a high-resolution
($R\!\sim\!28{,}000$) southern optical survey using the HERMES spectrograph on
the Anglo-Australian Telescope, designed for Galactic archaeology and
delivering up to $\sim$30 elemental abundances per star. We use the
continuum-normalized spectra of \cite{galah:dr4} over $4700$--$7900$\,\AA,
concatenated across the four HERMES arms onto a $14{,}420$-point grid, and learning, in addition to basic stellar parameters and $v\sin i$, also $11$ abundance elements ( $[\mathrm{Mg/Fe}]$, $[\mathrm{Si/Fe}]$, $[\mathrm{Ca/Fe}]$, $[\mathrm{Ti/Fe}]$,
  $[\mathrm{Na/Fe}]$, $[\mathrm{Al/Fe}]$, $[\mathrm{Mn/Fe}]$, $[\mathrm{Ni/Fe}]$,
  $[\mathrm{Cr/Fe}]$, $[\mathrm{Ba/Fe}]$, $[\mathrm{Y/Fe}]$). GALAH's
southern footprint is disjoint from \textit{Kepler}'s. 

\paragraph{\textit{Gaia}.} \textit{Gaia} \citep{gaia:mission} provides
all-sky astrometry and low-resolution spectrophotometry, and serves here as the
universal hub. We use two products from DR3 \citep{gaia:dr3}: (i) The BP/RP
(XP) \emph{sampled} spectra \citep{gaia:xp}: $340$ flux points spanning
$330$--$1050$\,nm, scaled per-spectrum with a robust (median/IQR) scaler and
deliberately \emph{not} continuum-normalized, so that the absolute
spectral-energy-distribution shape, which carries temperature and
extinction information, is preserved. (ii) The five-parameter astrometric
solution plus radial velocities, from which we derive $UVW$ space velocities via the equatorial-to-Galactic transformation of \citet{JohnsonSoderblom1987}.

\subsection{Cross-match}
\label{sec:data:crossmatch}

All surveys are cross-matched onto a common star identity keyed by the
\textit{Gaia}~DR3 \texttt{source\_id}.
The master cross-match retains every star observed by at least two of the five
non-\textit{Gaia} surveys, yielding \textbf{1{,}613{,}142} observations of \textbf{1{,}386{,}210} unique stars. Table~\ref{tab:crossmatch} lists the
per-survey census and the pairwise overlaps. The pair frequencies span roughly
two orders of magnitude---from TESS$\times$LAMOST (935{,}262 stars) down to
\textit{Kepler}$\times$APOGEE (20{,}712). Notably, \textit{Kepler} and GALAH have
\emph{zero} common stars (northern vs.\ southern footprints): these two
modalities are never observed together and can only be related transitively,
through the shared latent space. Counting the number of
non-\textit{Gaia} surveys per star, 1{,}225{,}913 stars have exactly two,
143{,}035 have three, and 17{,}262 have four; no star is covered by all five.

\begin{table}
\centering
\caption{Survey census and pairwise overlaps in the cross-match master (unique
stars; \textit{Gaia} is the anchor and covers essentially all
1{,}386{,}210 stars). The diagonal gives each survey's total; off-diagonal
entries give the number of stars observed by both surveys.}
\label{tab:crossmatch}
\begin{tabular}{lrrrrr}
\toprule
 & \textit{Kepler} & TESS & LAMOST & APOGEE & GALAH \\
\midrule
\textit{Kepler} & 91{,}213 & 69{,}701 & 66{,}615 & 20{,}712 & 0 \\
TESS   &  & 1{,}247{,}706 & 935{,}262 & 226{,}347 & 188{,}513 \\
LAMOST &  &  & 1{,}053{,}269 & 135{,}429 & 60{,}388 \\
APOGEE &  &  &  & 310{,}347 & 55{,}623 \\
GALAH  &  &  &  &  & 247{,}444 \\
\bottomrule
\end{tabular}
\end{table}

\begin{table}[t]
\centering
\small
\setlength{\tabcolsep}{4pt}
\begin{tabular}{llrlr}
\toprule
Modality & Architecture & Params & Input Length & $d_i$ \\
\midrule
\emph{Gaia} XP & Spectral ViT & 38.5\,M & $340$ & 512  \\
LAMOST & Spectral ViT & 615.3\,M & $4096$ & 2048 \\
APOGEE & Spectral ViT & 154.2\,M & $8576$ & 1024 \\
GALAH & Spectral ViT & 154.4\,M & $14{,}420$ & 1024 \\
\emph{Kepler} & Dual-view LC & 129.6\,M & $6\times21{,}600$ & 768 \\
TESS & Dual-branch LC & 139.5\,M & $4\times1000$ & 512 \\
\bottomrule
\end{tabular}
\caption{The six frozen unimodal encoders. Input Length is the number of flux bins (spectra) or channels and time samples (light curves). $d_i$ is the feature
dimensionality. }
\label{tab:encoders}
\end{table}

\subsection{Training}\label{sec:training}
Pre-training was done on a mixture of H100 and B200 GPUs. Each encoder was trained until convergence, using early stopping criteria. Training times vary between several hundred and several thousand GPU-hours. Training the alignment model is fast: of the order of $10$ B200 GPU-hours. 
The alignment model is trained on a fixed stratified subset that deliberately reserves
half its budget for single-modality stars, so the shared space is shaped by the
long unimodal tail and not only by the well-covered multimodal core. Multimodal stars are weighted by their rarest constituent modality pair.
The invariance and variance terms in VICReg were given a weight of $25$. All other loss terms had a weight of $1$.

\subsection{Why a continuity penalty produces a Fourier geometry}
\label{app:fourier}
We sketch why minimising the smoothness penalty organises the shared
coordinates into sinusoidal modes. The argument motivates the observed
geometry rather than proving a theorem about the full objective; idealisations
are stated as we go.

\paragraph{The unnormalised penalty carries the density squared.}
Write $s(c)$ for the shared embedding at parameter $c$. An unnormalised graph
penalty $\tfrac{1}{n^2}\sum_{a\neq b} w_{ab}\lVert s_a-s_b\rVert^2$ is a sum
over \emph{pairs} of stars drawn i.i.d.\ from the sampling density $p$, so its
expectation integrates against $p(x)p(y)$. Expanding at small bandwidth,
\begin{equation}
\frac{1}{n^2}\sum_{a\neq b} w_{ab}\lVert s_a-s_b\rVert^2
\;\longrightarrow\;
\varepsilon^{2}\!\int \lVert s'(c)\rVert^2\, p(c)^{2}\, dc ,
\end{equation}
one factor of $p$ per summation index. More generally, dividing the kernel by
$q_a^{\alpha}q_b^{\alpha}$ (with $q_a\!\propto\!p(c_a)$ the degree) gives a
continuum limit weighted by $p^{\,2-2\alpha}$ \citep{Coifman2006_diffusionmaps}.
A global normaliser such as $\sum w$ cannot change this exponent; only the
per-node division can.

\paragraph{Normalizing the density}
Taking $\alpha=1$, i.e.\ $\tilde{w}_{ab}=w_{ab}/(q_aq_b)$, cancels the density
to leading order and leaves the unweighted Dirichlet energy
$E[s]=\int\lVert s'(c)\rVert^2\,dc$. Its Euler--Lagrange operator, under
Neumann boundary conditions, is the ordinary Laplacian $-d^2/dc^2$.

\paragraph{The scale constraint carries the density too.}
The energy alone is minimised by a constant; a scale must be fixed, and this
is done by the VICReg variance and covariance terms. But these are batch
averages, hence estimates of integrals against $p$: the variance constraint
is ``unit variance under $p$'', which reintroduces the density through the
measure and warps the minimisers by a factor $\propto\sqrt{p}$. We therefore
apply the same $\alpha=1$ correction a second time, importance-weighting every
variance and covariance moment by $1/q_a\!\propto\!1/p(c_a)$ before the
(unchanged) hinge and off-diagonal penalties. Under a quadratic-constraint
idealisation the effective constraint becomes ``unit variance under Lebesgue
measure'', and the Rayleigh--Ritz problem is the density-free
\begin{equation}
-s_n'' = \lambda_n\, s_n .
\end{equation}
Normalising the kernel alone leaves the $\sqrt{p}$ warp; normalising both is
what makes the geometry density-free. The invariance term is left unweighted:
it compares two modalities of the same star, a per-pair quantity with no
density to cancel.

\paragraph{Fourier modes and their robustness.}
On $[0,L]$ with Neumann boundaries the eigenfunctions are
$s_n(c)=\cos(n\pi c/L)$, $\lambda_n=(n\pi/L)^2$: the observed ladder of
increasingly oscillatory coordinates. Even where the correction is imperfect,
the operator remains of Sturm--Liouville type, whose $n$-th eigenfunction has
exactly $n$ nodes, so the node ladder is preserved regardless of residual
density weighting. This robustness is why the ladder survives approximate
stationarity.

\paragraph{Why two coordinates degrade.}
The one-dimensional argument does not extend directly to two dimensions. The
distribution in $(T_{\mathrm{eff}},\log g)$ is neither translation invariant
nor defined on a rectangular domain, so higher eigenfunctions distort and mix
through near-degenerate eigenvalues. The lowest-frequency mode is far more
robust, explaining why only the fundamental remains clean in the
two-dimensional analysis.

\paragraph{Bandwidth.} $\varepsilon$ is fixed (chosen for ${\sim}10$ graph
neighbours per batch); edges are retained within a $3\varepsilon$ ball.
Optimising the bandwidth, including per-coordinate, is left to future work.

\clearpage
\subsection{Additional Results}

\begin{figure}[!ht]
  \centering
  \includegraphics[width=\textwidth]{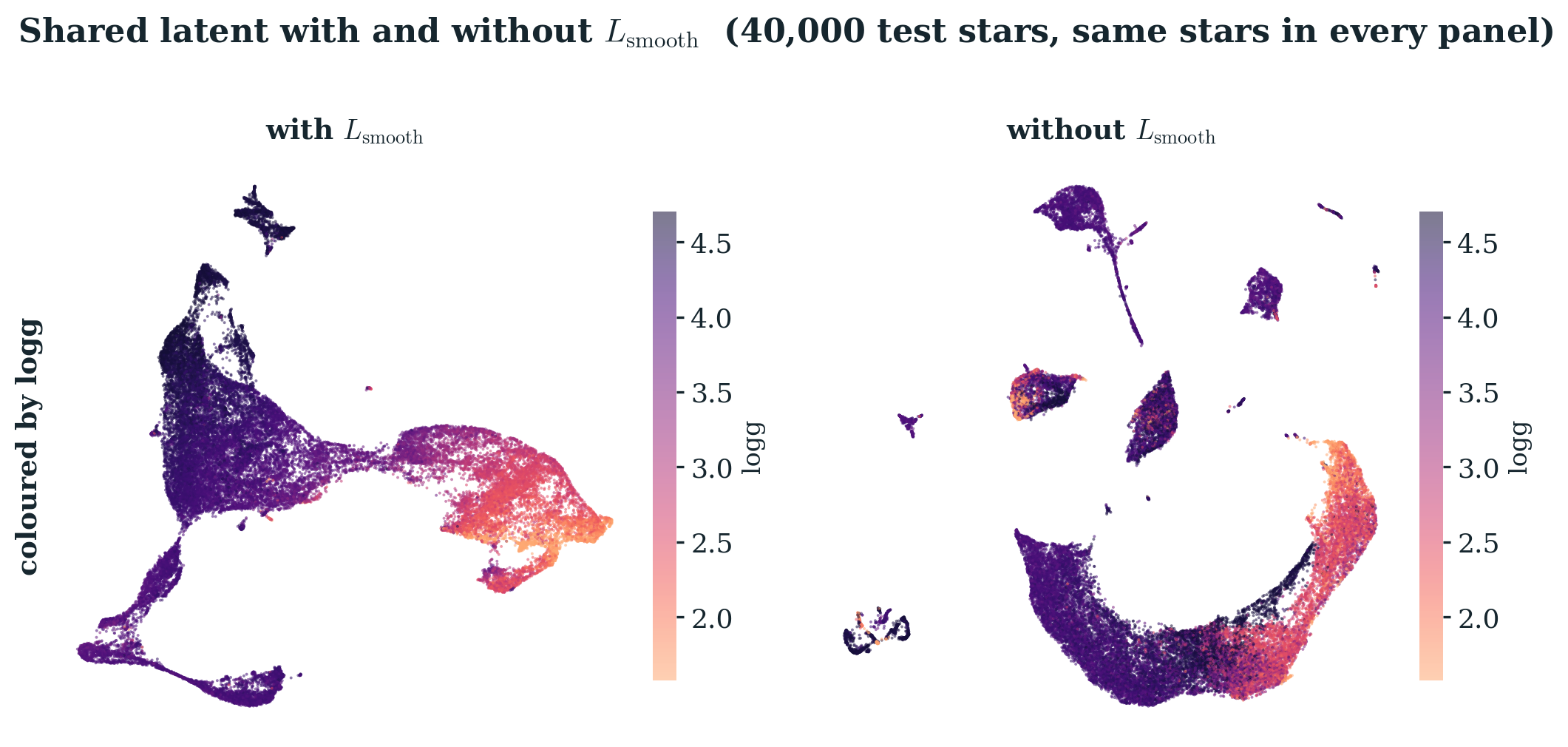}
  \caption{Effect of $\mathcal{L}_{\mathrm{smooth}}$ on the latent
    space. \textbf{Left:} UMAP of the shared latent space with
    $\mathcal{L}_{\mathrm{smooth}}$. \textbf{Right:} the same model
    without it. The prior yields a more continuous latent space with a
    clean $\log g$ gradient.}
  \label{fig:umap_smothnes}
\end{figure}

\subsection{Marginal shape}\label{app:arcsine}
A Fourier geometry constrains the feature distribution: a direction close to
$\cos(k\pi c)$ should be bimodal across stars (the arcsine law). We test this
per principal direction on frozen embeddings, calibrated against a synthetic
Fourier manifold (\Cref{tab:arcsine_cross}); per-block coverage is in
\Cref{tab:arcsine_blocks}. Past the third direction every model is heavy-tailed
rather than Gaussian, so ``arcsine head over a Gaussian bulk'' is not the
geometry these encoders produce.

\begin{table}[t]
\centering
\small
\setlength{\tabcolsep}{4pt}
\caption{Graph controls: linear-probe $R^2$ on the shared token
($n_{\mathrm{train}}=1000$, 30 paired draws) and $R^2_{\mathrm{Toep}}$
along $T_{\mathrm{eff}}$ on $11{,}347$ clean main-sequence singles
with LAMOST, APOGEE and \emph{Gaia}, fixed so that survey coverage
cannot drive the comparison. All arms share data, seed and recipe;
only the Laplacian's graph differs. The two FAME rows are the same
recipe trained twice and set the noise level ($\lesssim 0.03$ in
probe $R^2$, $0.04$ in $R^2_{\mathrm{Toep}}$). One training seed per
arm; best checkpoints fall at different epochs (3 to 79).}
\label{tab:smoothctl}
\begin{tabular}{lccccccc}
\toprule
Graph & $T_{\mathrm{eff}}$ & $\log g$ & [Fe/H] & Age & Act. & $v\sin i$ & $R^2_{\mathrm{Toep}}$ \\
\midrule
Physical (rerun)   & 0.872 & 0.873 & 0.530 & 0.557 & 0.165 & 0.481 & 0.852 \\
Physical (paper)   & 0.860 & 0.875 & 0.502 & 0.562 & 0.167 & 0.509 & 0.889 \\
No Laplacian       & 0.781 & 0.762 & 0.435 & 0.490 & 0.123 & 0.474 & 0.759 \\
Permuted           & 0.548 & 0.449 & 0.180 & 0.302 & 0.005 & 0.399 & 0.394 \\
Latent self-graph  & 0.705 & 0.504 & 0.204 & 0.448 & 0.109 & 0.442 & 0.775 \\
Input $z_i$        & 0.826 & 0.869 & 0.326 & 0.478 & 0.124 & 0.486 & 0.846 \\
Physical, $\alpha{=}0$ & 0.812 & 0.813 & 0.147 & 0.508 & 0.103 & 0.468 & 0.909 \\
Physical, warped   & 0.879 & 0.889 & 0.442 & 0.555 & 0.158 & 0.427 & 0.861 \\
\bottomrule
\end{tabular}
\end{table}

\begin{table}[t]
\centering
\small
\setlength{\tabcolsep}{4pt}
\caption{Marginal shape by survey block: excess kurtosis of PC1
($\gamma_2$; pure 1-D arcsine limit $-1.5$, Gaussian $0$), with
\emph{strong depth} as a subscript: the number of leading
contiguous PCs with $\gamma_2 < -1$. Bold marks
$\gamma_2(\mathrm{PC1}) < -1$. Star sets differ per block ($n$ in
the header row), so this table reports \emph{coverage}, not
cross-arm magnitude; for the latter see \Cref{tab:arcsine_cross}.
\emph{Kepler} is the minimum stratum ($n=1{,}819$) and its
verdict is the least secure. $\mathcal{L}_{\mathrm{smooth}}$ is the
density-normalised ($\alpha{=}1$) form throughout: a fixed-$\epsilon$
kernel divided by $q_a q_b$, with the VICReg variance and
covariance statistics importance-weighted by $1/q$. The prior takes
the signature from three blocks to six and tightens it onto the
arcsine value; without the prior the same architecture retains it in
four blocks, marginally, and loses TESS and GALAH.}
\label{tab:arcsine_blocks}
\begin{tabular}{lcccccc}
\toprule
& \emph{Kepler} & LAMOST & APOGEE & TESS & \emph{Gaia}~XP & GALAH \\
$n$ & 1{,}819 & 14{,}161 & 20{,}000 & 20{,}000 & 20{,}000 & 20{,}000 \\
\midrule
Raw $z_i$
 & $+449.38_{0}$ & $\mathbf{-1.16}_{2}$ & $\mathbf{-1.56}_{1}$ & $+22.44_{0}$ & $+4.21_{0}$ & $\mathbf{-1.69}_{1}$ \\
Vanilla VICReg
 & $-0.95_{0}$ & $-0.19_{0}$ & $-0.24_{0}$ & $-0.05_{0}$ & $-0.31_{0}$ & $+0.34_{0}$ \\
Shared, no $\mathcal{L}_{\mathrm{smooth}}$
 & $\mathbf{-1.26}_{1}$ & $\mathbf{-1.13}_{2}$ & $\mathbf{-1.00}_{2}$ & $+0.04_{0}$ & $\mathbf{-1.07}_{2}$ & $+0.82_{0}$ \\
Shared, with $\mathcal{L}_{\mathrm{smooth}}$
 & $\mathbf{-1.40}_{2}$ & $\mathbf{-1.37}_{2}$ & $\mathbf{-1.50}_{2}$ & $\mathbf{-1.33}_{2}$ & $\mathbf{-1.38}_{2}$ & $\mathbf{-1.54}_{1}$ \\
Residual, with $\mathcal{L}_{\mathrm{smooth}}$
 & $+446.25_{0}$ & $\mathbf{-1.10}_{2}$ & $-0.89_{0}$ & $+5.73_{0}$ & $+3.35_{0}$ & $\mathbf{-1.64}_{1}$ \\
\bottomrule
\end{tabular}
\end{table}

\begin{table*}[t]
\centering
\small
\caption{Common star set and calibration. \emph{Upper block:} tier
\texttt{cross} (clean main-sequence stars with both LAMOST and APOGEE present,
$n=10{,}214$; \emph{pooled} is the modality-averaged token on the largest exact
availability cell, $n=4{,}584$). Identical stars in every row, so this is the
only cross-arm comparison we make. \emph{Lower block:} synthetic manifolds
pushed through the identical pipeline at the measured signal fraction ($0.899$).
The density-matched control is the honest reference: a \emph{perfect} Fourier
manifold at this sample's own $T_{\mathrm{eff}}$ density reaches strong depth
$2$, so the achievable depth here is two to three directions.
$\mathcal{L}_{\mathrm{smooth}}$ is the density-normalised ($\alpha{=}1$) form
throughout. Single seed, $\varepsilon=0.0139$.}
\label{tab:arcsine_cross}
\begin{tabular}{ll rr c rr}
\toprule
Arm & Panel & $\gamma_2$(PC1) & $\gamma_2$(PC2) & Strong depth
    & Med.\ PC1--8 & Bulk \\
\midrule
Raw $z_i$
  & LAMOST & $-1.38$ & $-1.05$ & 2 & $+0.10$ & 0.00 \\
  & APOGEE & $-1.53$ & $-0.45$ & 1 & $+1.04$ & 0.00 \\
\addlinespace
Vanilla VICReg
  & LAMOST & $-0.29$ & $-0.32$ & 0 & $-0.02$ & 0.00 \\
  & APOGEE & $-0.42$ & $-0.52$ & 0 & $-0.08$ & 0.00 \\
  & pooled & $-0.48$ & $-0.60$ & 0 & $-0.08$ & 0.00 \\
\addlinespace
Shared, no $\mathcal{L}_{\mathrm{smooth}}$
  & LAMOST & $-0.86$ & $-1.10$ & 0 & $+1.52$ & 0.00 \\
  & APOGEE & $-0.84$ & $-1.11$ & 0 & $+1.52$ & 0.00 \\
  & pooled & $-0.22$ & $-1.21$ & 0 & $+1.30$ & 0.00 \\
\addlinespace
Shared, with $\mathcal{L}_{\mathrm{smooth}}$
  & LAMOST & $\mathbf{-1.54}$ & $-1.13$ & 2 & $+2.73$ & 0.00 \\
  & APOGEE & $\mathbf{-1.54}$ & $-1.13$ & 2 & $+2.74$ & 0.00 \\
  & pooled & $\mathbf{-1.58}$ & $-1.09$ & 2 & $+2.05$ & 0.00 \\
\addlinespace
Residual, with $\mathcal{L}_{\mathrm{smooth}}$
  & LAMOST & $\mathbf{-1.37}$ & $-0.78$ & 1 & $+1.73$ & 0.00 \\
  & APOGEE & $-0.88$ & $-0.65$ & 0 & $+0.52$ & 0.00 \\
\midrule
Control: Fourier, uniform coordinate & --- & $-1.49$ & $-1.43$ & 4 & $-0.73$ & 0.90 \\
Control: Fourier, measured density   & --- & $-1.51$ & $-1.32$ & 2 & $-0.08$ & 0.88 \\
Control: isotropic Gaussian          & --- & $-0.06$ & $-0.01$ & 0 & $+0.02$ & 0.81 \\
\bottomrule
\end{tabular}
\end{table*}
\end{document}